\RequirePackage{fix-cm}
\documentclass[twocolumn]{svjour3}          % twocolumn
\smartqed  % flush right qed marks, e.g. at end of proof
\usepackage{graphicx}
\usepackage{amsmath,amssymb,amsfonts}
\usepackage{multirow}
\usepackage[caption=false, font=footnotesize]{subfig}
\usepackage{booktabs}
\usepackage{color}
\begin{document}

\title{Multi-Scale RCNN Model for Financial Time-series Classification %\thanks{Grants or other notes
}


\author{Liu Guang        \and
	Wang Xiaojie \and
	Li Ruifan %etc.
}

%\author{}

\institute{Center for Intelligence of Science and Technology (CIST), Department of Computer Science, Beijing University of Posts and Telecommunications, Beijing, China
}
\date{Received: date / Accepted: date}

\maketitle

\begin{abstract}
Financial time-series classification (FTC) is extremely valuable for investment management. In past decades, it draws a lot of attention from a wide extent of research areas, especially Artificial Intelligence (AI). Existing researches majorly focused on exploring the effects of the Multi-Scale (MS) property or the Temporal Dependency (TD) within financial time-series. Unfortunately, most previous researches fail to combine these two properties effectively and often fall short of accuracy and profitability. To effectively combine and utilize both properties of financial time-series, we propose a Multi-Scale Temporal Dependent Recurrent Convolutional Neural Network (MSTD-RCNN) for FTC. In the proposed method, the MS features are simultaneously extracted by convolutional units to precisely describe the state of the financial market. Moreover, the TD and complementary across different scales are captured through a Recurrent Neural Network. The proposed method is evaluated on three financial time-series datasets which source from the Chinese stock market. Extensive experimental results indicate that our model achieves the state-of-the-art performance in trend classification and simulated trading, compared with classical and advanced baseline models.
	
\end{abstract}

\section{Introduction}
\label{sec:introduction}
Financial time-series classification (FTC) is highly important for investors. It emerges attention from wide research fields, especially the Artificial Intelligence (AI) ~\cite{kim2003financial}. The classical financial theory, Effective Market Hypothesis (EMH) ~\cite{b1}, suggests that every piece of information in the financial market affect the movements of the corresponding security price. Thus, numerous researches have investigated the impact of historical financial data for the future security price. Due to a large amount of constantly produced financial data, analyzing these data consumes massive labor work from the human expert. Consequently, the technologies which can automatically process these data have been widely explored~\cite{tsai2010combining,kara2011predicting,li2016empirical}. 

From the property of time-series, existing researches on FTC can be divided into the Multi-Scale (MS) -oriented methods and the Temporal Dependency (TD) -oriented methods.

For MS-oriented methods, existing researches focus on extracting the MS features from financial time-series. As we know, the high scale of financial time-series features reflects the trend information of the financial market in the long run, while the low scale financial time-series features embody the short-term trend information. The methods with only single-scale features neglect the information on other scales. Accordingly, these single-scale methods often fail to accurately describe the current state of time-series movement. Unsurprisingly, these methods tend to misjudge the category of financial time-series data. In order to describe financial time-series precisely, its MS-property should be considered. In the financial area, the MS-property of financial time-series has been extensively investigated~\cite{dacorogna1996changing}. By the similarity measured on multiple scales, the future price of given security can be estimated by finding similar history price sequence across different financial markets~\cite{papadimitriou2006optimal}. In the AI community, few studies have explored the MS-property of financial time-series. The most prior work, ScaleNet~\cite{geva1998scalenet}, decomposes the time-series into different scales by Wavelet transform. Then, it extracts features from each scale by different Neural networks to make a prediction. More recently, Cui et al.~\cite{cui2016multi} use Convolutional Neural Network (CNN) to improve the feature extraction ability. Although the above methods have achieved remarkable improvement compared to the methods only with single-scale features, these works overpass the TD within the financial time-series.

For TD-oriented methods, the non-linear models are often used due to the nonstationary of financial time-series. Most previous researches use classical models in modeling classification. For example, Kim~\cite{kim2003financial} uses the Support Vector Machine (SVM) to predict the stock price index. Compared to the Neural Network (NN), it achieves comparable results under their experiment setting. It is notable that these models are not specifically designed for modeling the TD. More recently, the deep learning models~\cite{deng2017hierarchical} are introduced to improve the feature extraction and representation from financial time-series. For instance, Recurrent Neural Network (RNN), which can handle TD effectively, is often used in this scenario~\cite{lin2017hybrid}. However, the above methods only use single-scale features and ignore the MS-property of financial time-series. Consequently, they are not capable to describe the current state of the financial market precisely.

The MS and TD property of financial time-series and the subtle relation between these properties make the FTC very challenging. Very few works have investigated the effect of employing both properties of financial time-series for FTC. Recently, State Frequency Memory (SFM)~\cite{hu2017state} integrate Long-Short Term Memory (LSTM) and Discrete Fourier Transform (DFT) to model the multiple frequency properties in stock price sequence. However, the DFT need pre-defined parameters which are very tricky and can not be learned automatically. In addition, the DFT is not a suitable choice for nonstationary financial time-series. Therefore, a new method for FTC which can effectively utilize both properties of financial time-series is needed.

To address the above problem, this paper proposes a Multi-Scale Temporal Dependent Recurrent Convolutional Neural Network (MSTD-RCNN) for financial time-series classification. The proposed model is an effective end-to-end model which can learn its parameters automatically. The major contributions of this paper are summarized as follows: 

\begin{itemize}
	\item[(1)] We propose a novel method for FTC which combine and utilize both MS and TD properties of financial time-series. The proposed method integrates CNN and RNN to handle two different properties in financial time-series.
	\item[(2)] MS features are extracted with CNN units from the single-scale input of financial time-series sequence. The parameters for each CNN units are learned automatically. There are no needs for tuning predefined parameters, which is critical for methods like DFT.
	\item[(3)] Different scales features are fused with an RNN. Benefited from its structure in handling TD, the RNN can explore and learn the dependency across different scales.
	\item[(4)] To evaluate MSTD-RCNN, we build three minute-level index price datasets, which are sourced from the China stock market. According to the financial time-series shares identical structure and properties, it is feasible to expand our methods to the global financial markets.
\end{itemize} 
The experimental results demonstrate that our model achieves superior performance compared to some classical and state-of-the-art baseline models in both financial time-series classification and simulated trading. 

The rest of this paper is organized as follows: Section \ref{sec:relatedwork} introduces Financial Time-series Prediction, the Multi-Scale (MS) property of time-series and GRU, Section \ref{sec:model} illustrates the formulation of FTC and architecture of MS-RCNN, Section \ref{sec:exp} gives the experimental settings, Section \ref{sec:results} describes the experimental results and analysis, Section \ref{sec:conc} shows the conclusions and future works.
\section{Related works}
\label{sec:relatedwork}
\subsection{Financial Time-series Prediction}

% For figure citations, please use "Fig" instead of "Figure".
Financial time-series prediction is essential for developing effective trading strategies in the financial market~\cite{lee1991inferring}. In past decades, it has attracted widespread attention from researchers of many areas, especially the Artificial Intelligence (AI) community~\cite{kim2003financial}. These researches mainly focus on a specifical market, e.g., the stock market~\cite{leung2000forecasting,saad1998comparative}, the foreign exchange market~\cite{frankel1990chartists,cheung2018exchange,das2017hybridized}, and the futures market~\cite{zirilli1996financial,kim2017intelligent}. Unsurprisingly, it is very challenging due to their irregular and noisy environment. 

From the perspective of the learning target, existing researches can be divided into the regression approaches and classification approaches. For the regression approaches, they treat this task as a regression problem~\cite{12,4}, aiming to predict the future value of financial time-series. While the classification-oriented approaches treat this as a classification problem~\cite{17,8}, focusing on financial time-series classification (FTC). 

In most cases, the classification approaches achieve higher profits than the regression ones~\cite{leung2000forecasting}. Accordingly, the effectiveness of various approaches in FTC has been widely explored~\cite{6,7,9,10}.

\subsection{Multi-scale of financial time-series}
The Multi-Scale (MS) property for time-series classification has been widely studied~\cite{peng1998multiple,papadimitriou2006optimal,stopar2018streamstory,yang2015deep,cui2016multi,wang2017time}. The concept of MS are often used for Computer Vision (CV) tasks~\cite{eigen2015predicting}, i.e., image object detection~\cite{cai2016unified}. An image is a sample formed by sampling the objects in the real world at a certain pixel level. Images in large-scale provide global features, images in small-scale provide local features. The MS of an image can provide more detailed information than single-scale features. 

Similar to images, time-series also typically have MS-property. Previous works mainly focus on predicting the future value or movement direction based on the assumption that the movement pattern of financial time-series will repeat itself. Thus, time-series similarity analysis approaches have been extensively investigated, i.e., discrete wavelet transform~\cite{geva1998scalenet}. Among these approaches, the use of MS-property is one of the key factors to measure the similarity between time-series sequences. Since the MS-property is very effective to characterize a time-series.

The way to analyze financial data draw more challenging due to their non-stationary characteristic and noisy environment in the financial market. Therefore, this paper focuses on predicting financial time-series movement direction by utilizing the MS-property.

\subsection{Temporal dependency of financial time-series}

Previous researches have explored the effectiveness of a method who can classify financial time-series based on their Temporal Dependency (TD). Traditionally, these researches can be divided into three categories: the feature-oriented methods, the model-oriented methods, and the integrated methods. 

For the feature-oriented methods, the key factor is to extract effective features from the financial time-series data. Statistical-based approaches, such as Principal Component Analysis (PCA)~\cite {tsai2010combining} and Information Gain (IG)~\cite{lee2009using}, are often used. These methods can help to improve the performance of a given model by removing the low relevant features. Some studies have introduced fuzzy logic to transform them into more expressive representations~\cite{guan2018novel,chang2008tsk,atsalakis2009forecasting}. Since these data are mainly numerical which are weakly expressive for category information. These researches transform the real value in a feature into a probability distribution over multiple categories, thereby improving the feature's expressive for category information. 

For the model-oriented methods, they focus on improving the fitting ability of the model. Traditionally, Support Vector Machine (SVM) and Neural Network (NN) are thought to be very effective for financial time-series classification~\cite{kara2011predicting}. Due to the excessive parameter size, they are easily over-fitting to the training set. As a result, Extreme Learning Machine (ELM)~\cite{ma2016selected} and Random Forest (RF)~\cite{patel2015predicting} is introduced for financial time-series classification. ELM can speed up training and improve generalization performance through randomly generated hidden layer units. RF ensembles multiple trees to achieve better prediction and generalization performance than a single model. In more recent, some pioneer researches have explored the effectiveness of deep learning models in financial time-series classification~\cite {liu2017foreign,akita2016deep,deng2017hierarchical}. Since deep learning models have many successful applications in Computer Vision (CV)~\cite{krizhevsky2012imagenet} and Natural Language Processing (NLP)~\cite{kim2014convolutional}. For instance, TreNet~\cite{lin2017hybrid} integrates Convolutional Neural Network (CNN) and Long-Short Term Memory (LSTM) for trend prediction. 

For the integrated methods, they often integrate multiple artificial intelligence or statistical-based techniques into a pipeline method for financial time-series classification. Some studies integrate the text classification~\cite{DeFortuny2014,Shynkevich2015} and sentiment analysis~\cite{12} in NLP with a classification model to determine the direction of the securities price movement. Kim and Han~\cite{kim2000genetic} have proposed feature selection methods based on Genetic Algorithm (GA) combined with a NN model to select useful features to predict the trend of stock price. Teixeira et al.~\cite{teixeira2010method} have used the technical indicators, which often are used in technical analysis, as the representation of financial data and feed them into the classification model for FTC. Durán-Rosal et al.~\cite{duran2017identifying} have used piecewise linear regression based turning points to segment the target sequence, and then use a NN to predict these points. In this work, we explore the effects of deep learning models integrate statical-based method (down-sampling) in FTC.

\section{Model}
\label{sec:model}
In this section, we provide the formal definition of the financial time-series classification. Then, we present the proposed MSTD-RCNN model.
\subsection{Problem formulation}
In this paper, we focus on classifying sequence of financial time-series data into different categories by their movement direction. The price of a given security in the financial market is often a sequence of univariable data sequence.  A financial time-series dataset is denoted as $ \mathbb{D} = \{(\mathbf{x}_i  ,y_i)\}_N $, where $ N $ is the number of samples in the dataset, $ \mathbf{x}_i  \in \mathbb{R}^{T}$ is the $ i $th sample with length $T$ and $  y_i \in \mathbb{R}$ is the corresponding label. Each sequence of time-series is denoted as $ \mathbf{x}=\{x_1,x_2,…,x_T  \} $, where $ x_j $ is the value at $ j $th time-step and $ T $ is the length of time steps.

As a result, FTC is to build a nonlinear map function from an input time-series $ \mathbf{x}_i $ to predict a class label $  y_i $ formula:
\begin{equation}
\centering
y_i = f(\mathbf{x}_i) ,\label{eq1}
\end{equation}
where $ f(\cdot) $ is the nonlinear function we aim to learn.

Financial Time-series Classification (FTC) emerge attentions from researchers of various fields. However, it is very challenging due to two major difficulties. Firstly, strategies/studies require Multi-Scale (MS) features to describe the state of the financial market. Secondly, the Temporal Dependency (TD) features of different scales are needed to be fused to make an accuracy classification. 

To address these problems for FTC, we propose a Multi-Scale Temporal Dependent Recurrent Convolutional Neural Network (MSTD-RCNN) model. The proposed model transform the input sequence into MS sequences, extract features from each scale, fuses these features and outputs the predicted category. Thus, the proposed model is an effective end-to-end model for FTC. 

\subsection{Model architecture}
\begin{figure*}
	\centering
	\includegraphics[scale=0.34]{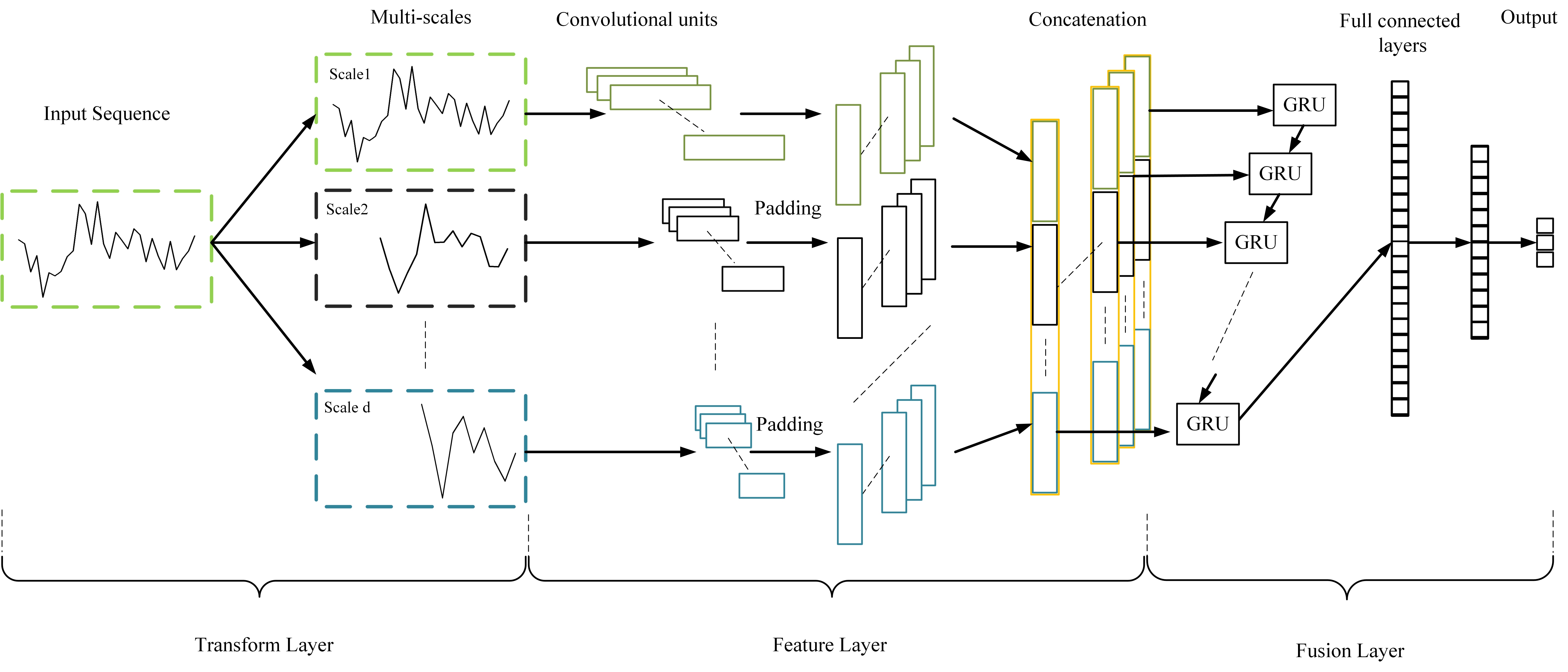}
	\caption{The architecture of MSTD-RCNN.\label{fig:fig1}}
	
\end{figure*}

The architecture of MSTD-RCNN is depicted in Fig.~\ref{fig:fig1}. Our model mainly has three components: the transform layer, the feature layer, and the fusion layer. The major functions of the three layers are described as follows:
\begin{itemize}
	
	\item[1] For the transform layer, the input sequence is transformed into MS sequences. Specifically, the down-sampling transformations in the time domain are used.
	
	\item[2] For the feature layer, different convolutional units are used to extract features from each scales. In this end, convolution units of different scales are independent of each other. The feature maps of the convolution output will be padded to the same length and then concatenated together.
	
	\item[3] For the fusion layer, we feed the padded and concatenated feature maps to the GRU. The output of the GRU passes through the fully connected layers and the softmax layer to produce the final output.
	
\end{itemize}
Thus, our MSTD-RCNN model is a complete end-to-end system where all parameters are jointly trained through backpropagation. 

\subsubsection{Transform layer} 

In this layer, the single-scale input sequence is transformed into multiple new sequences with different scales. Here, the down-sampling is used to generate sketches of financial data at different scales. This MS time-series is potentially crucial to the prediction quality for this task. Furthermore, they can complement each other. High scale features reflect slow trends and low scale features exhibit subtle changes in fast trends. 

Suppose there is a input sequence  $  \mathbf{x}=\{x_1,x_2,…,x_T   \} $, and the down-sampling rate is $d$. Then every $ d $th data points is keep in the new sequence $ \mathbf{x}^d=\{x_d,x_{2d},…,x_{md} \} $, where $ m=T/d $ is the length of sequence $ \mathbf{x}^d $. Through this method, multiple new sequences are generated with different down sampling rates, e.g., $ s = 1,2,…,d $. For simplify, we use $ \mathbf{X} $ to denote the generated sequences $ \{\mathbf{x}^1,\mathbf{x}^2,…,\mathbf{x}^d\} $.

\subsubsection{Feature layer}
This layer takes MS sequences as the input and outputs the concatenated features, which are extracted from each scale. It has two major components: the convolutional units and the concatenates operation.

\textit{Convolutional units.} The CNN units, which are often used as a feature extractor in Computer Vision~\cite{eigen2015predicting}, are used to extract feature maps from sequences with different scales. Specifically, 1-dimension CNN is used to process these newly generated sequences. These CNN units share the same filter size and number across all these sequences. Note that, with the same settings, higher scale sequence would get a larger receptive field than the original sequence. Through this means, each output of the convolution operation captures the features with a different receptive field from the original sequence. An advantage of this process is that by down-sampling the input sequence instead of increasing the filter size, we can greatly reduce the computation in the convolutional units.

Let $ \mathbf{x}^d $ to denote the $d$th scale time-series. The corresponding kernel weights $ \mathbf{w}_j^d \in \mathbb{R}^k $ is used to extract features from the input sequence. Here, $ k $ is the window size. For instance, the feature $ c_{j,i}^d $ is calculated by
\begin{equation}
c_{j,i}^d=f_a(\mathbf{w}_j^d * \mathbf{x}_{i:i+k-1}^d +b_j^d)\,.\label{eq2}
\end{equation}
Here, $*$ indicates the convolution operation, $ \mathbf{b}_j^d\in \mathbb{R}  $ is a bias term and $ f_a (\cdot) $ is a non-linear function such as the Rectified Linear Unit (ReLU). This filter is applied to the sequence $ \{\mathbf{x}_{(1:k)}^d,\mathbf{x}_{(2:k+1)}^d,…,\mathbf{x}_{(m-k+1:m)}^d\} $ to produce a feature map $ \mathbf{c}_j^d\in \mathbb{R}^{(m-k+1)}$ as follow
\begin{equation}
\mathbf{c}_j^d=[ c_{(j,1)}^d, c_{(j,2)}^d,…, c_{(j,m-k+1)}^d ]\,. \label{eq3}   
\end{equation}
%%跟pooling 的异同
Here, the pooling layers are not used. Since the transform layer does similar work as the pooling layers. The pooling layers increasing the receptive field~\cite{howard2017}. While the transform layer transforms the original sequence into different time-scales before the feature extraction. We believe the transformation before the feature extraction can achieve similar effects with pooling layers.\\
%%跟dilated convolutional的异同
%%跟多个卷积核的异同
\textit{Concatenation operation.} This operation concatenates the feature maps of different scales. Due to the different lengths of feature maps, padding is needed before concatenation. 

Since $ m=T/d $, the length of the feature map decreases with scale increasing. For the convenience of calculation, we unify the feature maps of different scales to the same length, that is, the feature map length $ T-k+ 1$ when $  d=1 $. We align the feature maps of other scales to $ T-k+1 $ length by zero-padding. For example, the alignment of the feature map for scale $d$ is as follow
\begin{equation}
\mathbf{a}_j^d=[z_1,z_2,…,z_{T-m},c_{j,1}^d,c_{(j,2)}^d,…,c_{(j,m-k+1)}^d]\,,\label{eq4}
\end{equation}
where $ \mathbf{a}_j^d \in \mathbb{R}^{T-k+1}$ is the padded feature map generated by jth kernel for $d$ scale and $ z_1,z_2,…,z_{(T-m)} $ are zeros sequence with length $ T-m $.

Next, the padded feature maps are concatenated into a feature matrix. The concatenating process is described as following
\begin{equation}
\mathbf{E}=[(\mathbf{a}_1^1)^T,\cdots,(\mathbf{a}_l^{1})^T,(\mathbf{a}_1^2)^T,\cdots, (\mathbf{a}_l^d)^T]^T,\label{eq5}
\end{equation}
where $ \mathbf{E} \in \mathbb{R}^{(d\times l) \times (T-k+1)} $ is the feature matrix with length $ T-k+1 $, $l$ is the number of convolutional kernels.

\subsubsection{Fusion layer} 
The fusion layer fuses the features from multi-scales and generates a prediction. The output of the feature layer is similar to the language model in Natural Language Processing, which has Temporal Dependency (TD) among each node. The major difference is that the sequences from different scales have different fields of view. To fuse these features, we need a model that captures this dependency and variety. The Recurrent Neural Networks (RNN) is often used as an encoder in Machine Learning Translation~\cite{cho2014learning}. It can capture the complex dependency in different languages. Hence, we use the RNN model to process the feature maps in this case.

Recurrent Neural Networks (RNN) have been successfully applied in machine translation~\cite{sutskever2014sequence}. The structure of RNNs are good at handling a variable-length sequence input by having a recurrent hidden state whose activation at each time is dependent on that of the previous time.

Similar to a traditional neural network, we can use a modified backpropagation algorithm Backpropagation Through Time (BPTT) to train an RNN~\cite{mozer1989focused}. Unfortunately, it is difficult to train RNN to capture long-term dependencies because the gradients tend to either vanish or explode~\cite{bengio1994learning}. Hochreiter and Schmidhuber~\cite{hochreiter1997long} proposed a long short-term memory (LSTM) unit and Cho et al.~\cite{cho2014learning} proposed a gated recurrent unit (GRU) to deal with the Problem effectively. To this end, we use GRU to process the feature matrix.

A Gated Recurrent Unit (GRU) makes each recurrent unit to adaptively capture dependencies of different time scales. The parameters can be updated by the following equations
\begin{equation}
r_t=\sigma(W_r x_t+U_r h_{t-1})\,,\label{eq6}
\end{equation}
\begin{equation}
z_t=\sigma(W_z x_t+U_z h_{t-1})\,,\label{eq7}
\end{equation}
\begin{equation}
h'_t=tanh(W_h x_t+U_h (r_t \odot h_{t-1} ))\,,\label{eq8}
\end{equation}
\begin{equation}
h_t=(1-z_t)h_{t-1}+z_t h'_t\,,\label{eq9}
\end{equation}
where $\sigma(\cdot)$ denotes the logistic sigmoid function, $\odot$ denotes the element-wise multiplication, $ r_t $  denotes the reset gate, $ z_t $ denotes the update gate and $ h'_t $ denotes the candidate hidden layer. In this paper, we apply the GRU as the feature summarize layer for stock trend prediction.

Given the feature vector $ \mathbf{e}_t$, the hidden states at the $t$th time-step can be calculated by
\begin{equation}
\mathbf{h}_t=f_{en} (\mathbf{e}_t,\mathbf{h}_{t-1};\boldsymbol{\theta}_{en} ) \,,\label{eq10}
\end{equation}                             
where $ \mathbf{h}_t\in \mathbb{R}^q $ is the hidden state of the encoder at time $ t $, $ q $ is the size of hidden state, $\mathbf{e}_t \in \mathbb{R}^{d\times l}$ is the $t$th column in the matrix $\mathbf{E}$, $f_{en} (\cdot) $ is a non-linear function, and $ \boldsymbol{\theta}_{en} $  is the parameters of encoder function. There are many choice for encode the sequence of numerical data. In this case, we use GRU as the non-linear function. The output of GRU is $ \mathbf{h}_{T-k+1} \in \mathbb{R}^q $ is deemed as the encoding of the multiple scales of input sequence.\\

The feature vector output by the GRU is passing through the multiple fully connected layers, and then a softmax activation layer to obtain a probability distribution of different classes. The softmax activation function is calculated as follow
\begin{equation}
p'_i =\frac{e^{\mathbf{o}_i}}{\sum_{j=1}^{C}e^{\mathbf{o}_j}}\,,\label{eq11}
\end{equation}
where $\mathbf{o}_i$ indicates the result of the $i$th output node, and $C$ is the number of categories.

The cross-entropy loss function is used to measure the difference between our predicted classification distribution $ p'_{t,i} $ and real distribution $ p_{t,i} $:
\begin{equation}
J_{\boldsymbol{\theta}} = - \frac{1}{N} \sum_{t=1}^{N} \sum_{j=1}^{C} p_{t,j} \log(p'_{t,j})\,,\label{eq12}
\end{equation}
where $ \boldsymbol{\theta} $ is represent all the parameter of the model, $ N $ is the total number of samples. 

\section{Experimental settings}
\label{sec:exp}
In this section, we first give the details of datasets. Then, we introduce the baseline models in comparative evaluation.  Last, the evaluation metrics are illustrated. 
\subsection{Datasets} 
We first describe the data source of the datasets. Then, we explain how to choose the threshold for the label and the window size for window sliding.

Three high-frequency stock index datasets are collected from the Chinese stock market. 
\begin{itemize}
	\item{SH000001:} Shanghai Stock Exchange (SSE) Composite Index. Prepared and published by SSE index is the authoritative statistical indices widely followed and used at home and abroad to measure the performance of China's securities market. 
	
	\item{SZ399005:} Shenzhen Stock Exchange Small \& Medium Enterprises (SME Boards) Price Index. SME play an important role in the economic and social development of China. They foster economic growth, generate employment and contribute to the development of a dynamic private sector.
	
	\item{SZ399006:} ChiNext Price Index is a NASDAQ-style board of the Shenzhen Stock Exchange. It aims to attract innovative and fast-growing enterprises, especially high-tech firms. Its listing standards are less stringent than those of the Main and SME Boards of the Shenzhen Stock Exchange.  
	
\end{itemize}
The data in the dataset begins on January 1, 2016, and ends on December 30, 2016. There are a total of 58,000 data points. The window slicing is applied for the data augmentation~\cite{cui2016multi}. There are 48,000 of data points are used as training sets, 5000 are used as verification sets, and 5000 are used as testing sets.\\
There are three categorical values, they are defined as follows 
\begin{equation}
y_t=\left\{\begin{matrix}
-1&  \Delta x_t \le -\delta\\
0& -\delta < \Delta x_t < \delta  \\ 
1&  \Delta x_t \ge \delta
\end{matrix}\right.\,.\label{eq13}
\end{equation}
Here,  $c=0$ means the price of the security in the next time-step is still, $ c=1 $ means the price is going upward and $ c=-1 $ means the price is moving downward, $ \delta $ is the threshold and $\Delta x_t$ is the change value compared to the previous time-step, it calculated by
\begin{equation}
\Delta x_t=x_t-x_{t-1}\,. \label{eq14}
\end{equation}

\begin{figure*}
	\centering
	\includegraphics[scale=0.5]{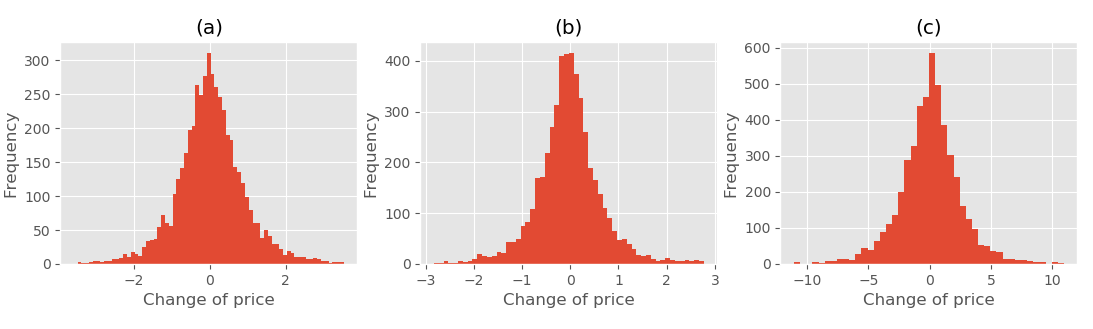}
	\caption{(a)The histogram of price change on development set of SH000001. (b)The histogram of price change on development set of SZ399006. (c)The histogram of price change on development set of SZ399005.\label{fig4}}
\end{figure*}

To select the threshold for each dataset, we analyzed the distribution of each dataset. As shown in Fig.~\ref{fig4} and Table~\ref{tab8}, the distribution of price change on the development set on each dataset are mostly clustered around zero. We choose the threshold which can make each category on the development set distribute equally. As a result, the threshold is set to 0.3, 0.2 and 0.8 for SH000001, SZ399006, and SZ399005. The distribution of each dataset is shown in Table~\ref{tab6}. 

To select the window size for sliding windows. We use the Random Forest(RF) to train on the training set and evaluate the development set under different window size setting. As shown in Table~\ref{tab7}, the window size $ T=30 $ makes the superior performance for RF. Therefore, the window size set to $30$.\\
	
\begin{table*}[ht]
	\centering
	\caption{Ratio of categories on each datasets. \label{tab6}}
	\setlength{\tabcolsep}{2mm}{\begin{tabular}{cccccccccc}
			\toprule
			&\multicolumn{3}{c}{SH000001}&\multicolumn{3}{c}{SZ399005}&\multicolumn{3}{c}{SZ399006}\\\midrule
			Category(trend)		& Train(\%)   & Dev(\%)  & Test(\%) & Train(\%)   & Dev(\%)  & Test(\%) & Train(\%)   & Dev(\%)  & Test(\%) \\\midrule
			Downward($\downarrow$) & 36.60    & 33.04   &  35.66 & 41.68    & 35.78   &  36.76 & 40.31   & 32.82   &  35.28 \\
			Still(-)               & 26.99    & 32.16   &  31.90 & 18.67    & 32.42   &  31.60 & 18.50   & 33.16   &  29.24  \\
			Upward($\uparrow$)     & 36.41    & 34.80   &  32.44 & 39.65    & 31.80   &  31.64 & 41.19   & 34.02   &  35.48 \\
			\bottomrule
	\end{tabular} }
\end{table*}

\begin{table*}[ht]
	\centering
	\caption{Window size on each development sets. \label{tab7}}
	\setlength{\tabcolsep}{6.3mm}{\begin{tabular}{ccccccc}
			\toprule
			&\multicolumn{2}{c}{SH000001}&\multicolumn{2}{c}{SZ399005}&\multicolumn{2}{c}{SZ399006}\\\midrule
			window size	& Acc   & F1  & Acc   & F1 & Acc   & F1 \\\midrule
			10 & 0.5132 & 0.5123 & 0.4842 & 0.4718 & 0.5874 & 0.5826\\
			20 & 0.5212 & 0.5214 & 0.5018 & 0.4905 & 0.5974 & 0.5932\\
			30 & \textbf{0.5256} & \textbf{0.5252} & \textbf{0.5064 }& \textbf{0.4938} & \textbf{0.5996} & \textbf{0.5956}\\
			40 & 0.5210 & 0.5207 & 0.4856 & 0.4703 & 0.5950 & 0.5901\\
			50 & 0.5222 & 0.5212 & 0.4998 & 0.4842 & 0.5934 & 0.5877\\
			\bottomrule
	\end{tabular} }
	\footnotesize{ \\Bold numbers indicate the best results}
\end{table*}

\begin{table}[ht]
	\centering
	\caption{Statical features of each development sets. \label{tab8}}
	\setlength{\tabcolsep}{4mm}{\begin{tabular}{cccc}
			\toprule
			&SH000001&SZ399005&SZ399006\\\midrule
			Mean & 0.0368 & 0.0016 & 0.0081\\
			Std  & 1.1915 & 0.9429 & 3.6840 \\
			\bottomrule
	\end{tabular} }
\end{table}

In order to avoid excessive correlation between these datasets, we calculate the Pearson Correlation Coefficient (PCC) for the data of these three data sets. $PCC_{XY}=-1$ indicates that $X$ and $Y$ has a negative correlation. $PCC_{XY}=1$ indicates that $X$ and $Y$ has a positive correlation. $PCC_{XY}=0$ indicates that $X$ and $Y$ has no correlation. Table~\ref{tab1} lists the results of PCC, which indicate that there are no strong correlations ($<0.5$) between each pair of datasets.

\begin{table}[ht]
	\centering
	\caption{PCC results between each pair of datasets. \label{tab1}}
	\setlength{\tabcolsep}{3.5mm}{\begin{tabular}{cccc}
			\toprule
			&SH000001&SZ399005&SZ399006\\\midrule
			SH000001 & 1.00    & 0.58   &  0.42 \\
			SZ399005 & -      & 1.00    &  0.42 \\
			SZ399006 & -      & -      &   1.00 \\
			\bottomrule
	\end{tabular} }
\end{table}
\subsection{Baselines}
There are six baseline models are used. Firstly, two classical models in FTC are given. Then, four advanced models for FTC are illustrated.
\begin{itemize}
	\item{Support Vector Machine (SVM)~\cite{kim2003financial}.} It projects the input data into a higher dimensional space by the kernel function and separates different classes of data using a hyperplane. The trade-off between margin and misclassification errors is controlled by the regularization parameter. 
	
	\item{Random Forest (RF)~\cite{kara2011predicting}.} It belongs to the category of ensemble learning algorithms. It uses the decision tree as the base learner of the ensemble. The idea of ensemble learning is that a single classifier is not sufficient for determining the class of test data. After the creation of n trees, when testing data is used, the decision on which the majority of trees come up with is considered as the final output. This also avoids the problem of over-fitting.
	
	\item{Fuzzy Deep Neural Network (FDNN)~\cite{deng2017hierarchical}.} FDNN uses fuzzy-neural layers and fully connected layers to learn the fuzzy representation and neural representation separately. Then, these two representations are fused by a two-layer fully connected layer. The fused representations are fed to a softmax activation to get the trend to predict results.
	
	\item{TreNet~\cite{lin2017hybrid}.} TreNet hybrids LSTM and CNN for stock trend classification. Firstly, LSTM learning the dependencies in historical trend sequence,  and CNN learning the local features from raw data of time-series. Then, these extracted features fused by a fully connected layer to generate a prediction.
	
	\item{State-Frequency Memory Recurrent Neural Networks (SFM) \cite{hu2017state}.} It allows separating dynamic patterns across different frequency components and their impacts on modeling the temporal contexts of input sequences. By jointly decomposing memorized dynamics into state frequency components, the SFM is able to offer a fine-grained analysis of temporal sequences by capturing the dependency of uncovered patterns in both time and frequency domains.
	
	\item{Multi-Scale CNN (MS-CNN)~\cite{cui2016multi}.} MS-CNN uses different convolutional units to extract features from each time-scale of data. Then, these features are fused by a two-layer fully connected layers. In most cases, this model achieves better performance than the regular convolutional neural network in time-series classification.
\end{itemize}
The parameters of our model are selected by the performance on the validation set. The maximum epoch is set to 100. The model is trained by Adam optimization algorithm with the learning rate 0.0005. The batch size is set to 32. There are $l=3$ time-scales. Convolution unit for each time-scale has 16 filters, the number of hidden units in GRU is set to 48 ($3 \times 16$).\\

\subsection{Evaluation metrics}
Accuracy, F-score(F1) and Confusion Matrix(CM) are used as the classification metrics to evaluate the models. And the accumulated profit is used to evaluate the profitability of the models. \\
Accuracy and F1 are calculated based on Confusion Matrix which has four components: True Positive(TP), True Negative(TN), False Positive(FP) and False Negative(FN). CM shows for each pair of classes $\langle c_1,c_2\rangle$, how many samples from $c_1$ were incorrectly assigned to $c_2$. \\
Accuracy is the rate of correct prediction and is calculated as the formula in Equation~\ref{eq15}. Equation~\ref{eq15}.
\begin{equation}
acc=\frac{TP+TN}{N} \,.\label{eq15}
\end{equation}
Here, $ N $ is the total number of samples in dataset. We next explain the weighted average of  F1. The calculation of F1 is displayed in Equation~\ref{eq16}.
\begin{equation}
f1=\frac{2\times P \times R}{P+R}\,,\label{eq16}
\end{equation}
where R is the recall and P is the precise, which are calculated as follows:
\begin{equation}
P=  \frac{TP}{TP+FP}\,,\label{eq17}
\end{equation}
\begin{equation}
R=  \frac{TP}{TP+FN}\,.\label{eq18}
\end{equation}

The simulated trading algorithm is calculated based on the predicted result $c'_t$, the real trend $c_t$ and the change value of index $\Delta x_t$. For each trading signal generated by the model, we will execute the buy-in or sell-out one unit of security. For the upward and downward categories, we will make a profit if the prediction is correct, and if it is wrong, we will suffer losses. For the still category, the change value $\Delta x_t=0$ is set to zero. The transaction cost is set to zero. The accumulated profit  $P.$ is calculated by
\begin{equation}
P. = \sum_{t=1}^{N} I(c'_t,c_t)\times \Delta x_t\,.\label{eq19}
\end{equation}		
Here, $P.$ indicates the profit representing the change points, $I(c'_t,c_t)$ is an indicator function, which equals 1 when the $c'_t = c_t$, otherwise 0.

\section{Results and Analysis}
\label{sec:results}
First, the model’s performance is compared with the baseline models on three datasets. Then, the effects of the feature layer in extracting Multi-Scale (MS) features are analyzed. Third, the effects of the fusion layer in capturing Temporal Dependent (TD) are analyzed. Next, the profitability of models is evaluated by simulated trading. Last, the reason for driving improvement in profitability is analyzed through the confusion matrix.

\subsection{Comprehensive evaluation}
Financial time-series classification is a challenging task and a minor improvement usually leads to large potential profits~\cite{4}. To demonstrate the effectiveness of our MSTD-RCNN model, we compare it against the six baseline models on three datasets. The results are listed in Table~\ref{tab2} The t-test results between our model and other models are listed in Table~\ref{tab3}. Examining the experimental results, we reach the following conclusions.

\begin{itemize}
	\item[1] Our model achieves the best performance in both accuracy and F1. From the perspective of accuracy, our model achieves the best results in all three datasets. Especially, MSTD-RCNN rises the accuracy of 3.07\%, 3.00\%, and 2.13\% higher than the best baseline models on SH000001, SZ399005, and SZ399006. From the perspective of F1, our model also achieves the best performance on these three datasets. Especially, MSTD-RCNN has 4.14\%, 2.21\%, and 2.86\% improvement compared to the best baseline models on SH000001, SZ399005, and SZ399006. In addition, the t-test results suggest that the results of our model are significantly different from the ones of other models.
	
	\item[2] Our MSTD-RCNN model can effectively extract MS features from financial time-series. First of all, all models share the same single-scale input. Only SFM, MS-CNN and MSTD-RCNN are designed to utilize the MS-property of time-series. As a result, these models achieve a higher level of accuracy and F1 than other baseline models in most cases. It can be concluded that the FTC models, which can utilize MS-property, are more effective than the single-scale ones. Secondly, our model achieves the best accuracy and F1 performance among these three models. That indicates our MSTD-RCNN is more effective in extract MS features than the other two models. 
	
	\item[3] Our MSTD-RCNN model is very effective in capturing Temporal Dependency (TD) within financial time-series. MSTD-RCNN has a significantly higher level of classification performance than MS-CNN. These two models have a similar structure. Especially, they both transform input single-scale sequence into MS sequences and then use CNN to extract MS features. For fuse features, MSTD-RCNN uses the GRU and the MS-CNN uses NN. We can conclude that the performance improvement is likely due to the efficiency of our MSTD-RCNN in capturing TD.
	
\end{itemize}
\begin{table*}[!h]
	\centering
	\caption{Results on the three datasets.\label{tab2}}
	\setlength{\tabcolsep}{6.8mm}{\begin{tabular}{ccccccc}
			\toprule
			\multirow{2}{*}{Model} &\multicolumn{2}{c}{SH000001}&\multicolumn{2}{c}{SZ399005}&\multicolumn{2}{c}{SZ399006}\\\cline{2-7}
			\specialrule{0em}{2.3pt}{1pt}
			& ACC & F1 & ACC & F1 & ACC & F1 \\\midrule
			SVM&0.5150&0.5181&0.5038&0.4960&0.5620&0.5357\\
			RF&0.5230&0.5196&0.5116&0.5054&0.5732&0.5654\\
			TreNet&0.5238&0.5250&0.5120&0.5134&0.5964&0.5857\\
			FDNN&0.5232&0.5245&0.5122 &0.5035&0.5930&0.5510\\
			SFM&0.5296&0.5227&0.5254&0.5232&0.5960&0.5740\\
			MS-CNN&0.5334&0.5287&0.5198&0.5201&0.6006&0.5954\\
			MSTD-RCNN&\textbf{0.5498}&\textbf{0.5506}&\textbf{0.5454}&\textbf{0.5516}&\textbf{0.6134}&\textbf{0.6124}\\\bottomrule
	\end{tabular}}
	\footnotesize{ \\Bold numbers indicate the best results}
\end{table*}
\begin{table*}
	\centering
	\caption{The t-test results between MSTD-RCNN and other baseline models.\label{tab3}}
	\setlength{\tabcolsep}{5.8mm}{\begin{tabular}{ccccccc}
		\toprule
		& SVM & RF & TreNet & FDNN & SFM & MS-CNN \\\midrule
		MSTD-RCNN ($\times 10^{-3}$)&\bf{0.8}*** &\bf{0.1}*** &\bf{0.2}*** &\bf{2.5}*** & \bf{0.5}*** & \bf{0.7}***\\\bottomrule
	\end{tabular}}
	\footnotesize{\\p-value $< 0.01$: ***, p-value $< 0.05$: **, p-value $< 0.1$: *}
\end{table*}

\subsection{Effects of multi-scale features}
To illustrate the effects of MSTD-RCNN in employing Multi-Scale (MS) property, we evaluate our MSTD-RCNN under different scale settings. Since MSTD-RCNN uses the convolutional unit to extract features from distinct scales. The effects on feature extraction can be evaluated by the classification performance when using different scale settings. Hence, our model is evaluated with scale settings as follows: $(1),(1,2),(1,2,3)$. $(1)$ indicates our model using original single-scale data. $(1,2)$ suggest our model using scale $1$ and $2$. $(1,2,3)$ denotes our model using three corresponding scales. 

Table~\ref{tab4} lists the classification results of MSTD-RCNN with different scale settings. The accuracy and F1 of MSTD-RCNN are rising with the increasing of scale number. The  $(1,2,3)$ MSTD-RCNN outperforms the $(1,2)$ MSTD-RCNN, and the $(1,2)$ MSTD-RCNN outperforms the $(1)$ MSTD-RCNN. These are mainly due to the effect of MS features. These features from different scales can complement each other.  Moreover, MSTD-RCNN achieves higher classification performance than baselines even if only using single-scale data. The $(1)$ MSTD-RCNN achieves a higher level of accuracy and F1 than baselines on three datasets. This is likely due to the effect of the convolutional units in feature extraction.

\begin{table*}[!h]
	\centering
	\caption{Effects of multi-scale features.\label{tab4}}
	\setlength{\tabcolsep}{6mm}{\begin{tabular}{ccccccc}
		\toprule
		\multirow{2}{*}{Model} &\multicolumn{2}{c}{SH000001}&\multicolumn{2}{c}{SZ399005}&\multicolumn{2}{c}{SZ399006}\\\cline{2-7}
		\specialrule{0em}{2.3pt}{1pt}
		& ACC & F1 & ACC & F1 & ACC & F1 \\ 
		\midrule
		$(1)$ MSTD-RCNN&0.5356&0.5373&0.5330&0.5210&0.6070&0.6040\\
		$(1,2)$ MSTD-RCNN&0.5454&0.5442&0.5414&0.5410&0.6122&0.6112\\
		$(1,2,3)$ MSTD-RCNN&\textbf{0.5498}&\textbf{0.5506}&\textbf{0.5454}&\textbf{0.5516}&\textbf{0.6134}&\textbf{0.6124}\\\bottomrule
	\end{tabular}}
	\footnotesize{ \\Bold numbers indicate the best results}
\end{table*}

\subsection{Effects of temporal dependency}
To show the effects of Temporal Dependency (TD), we compare the classification performance of MS-CNN and our model under different scale settings. MS-CNN and our MSTD-RCNN share similar structure in feature extraction. Hence, the differences in classification performance are majorly due to the different level of efficiency in capturing TD. 

\begin{figure*} 
	\centering
	\subfloat[\label{a}]{%
		\includegraphics[width=0.5\linewidth]{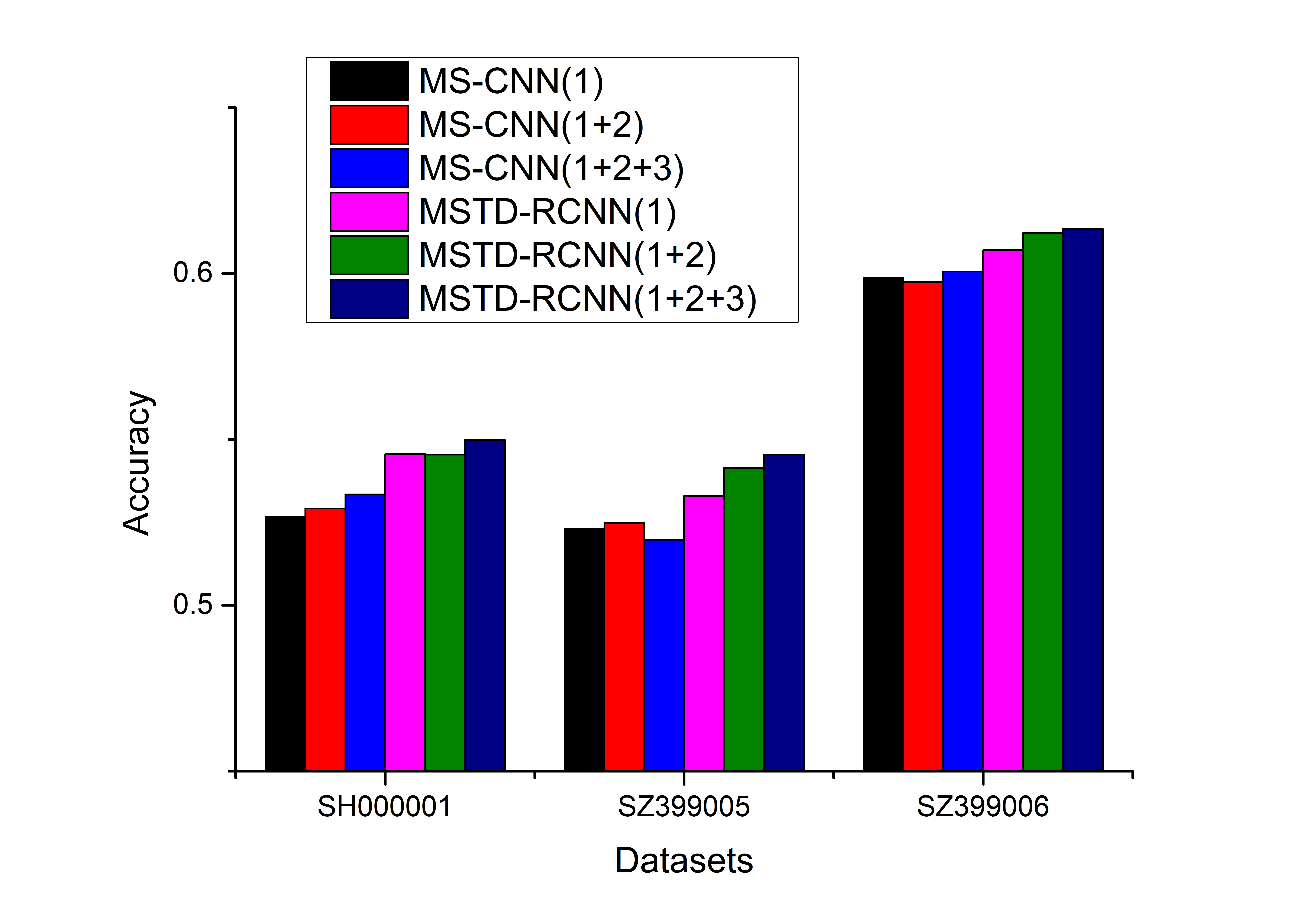}}
	\hfill
	\subfloat[\label{b}]{%
		\includegraphics[width=0.5\linewidth]{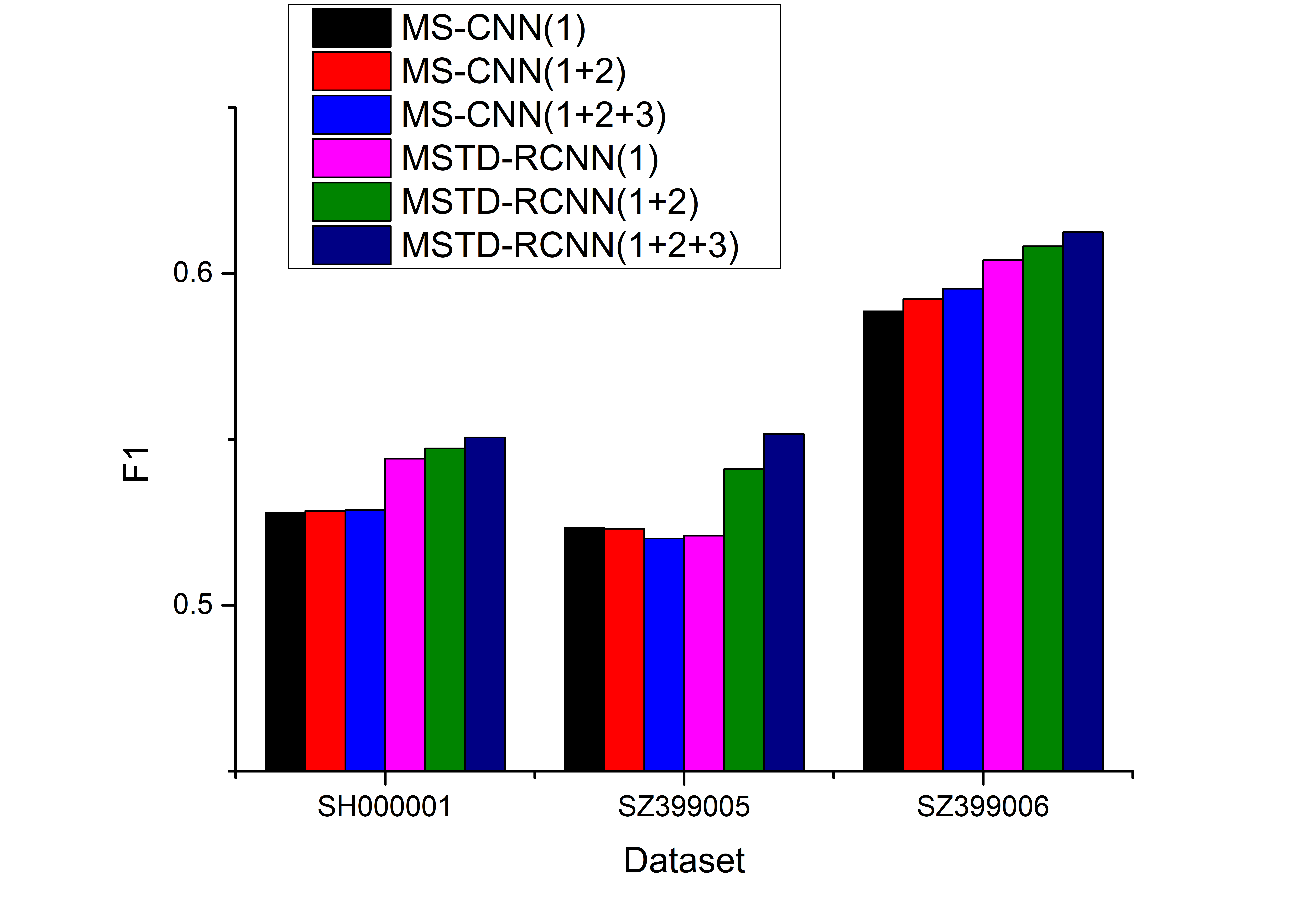}}
	\caption{(a)The accuracy results with different scales on three datasets. (b)The F1 results with different scales on three datasets.}
	\label{fig2} 
\end{figure*}

Fig.~\ref{fig2}.a shows the accuracy results on three datasets. Firstly, the performance of MSTD-RCNN is rising with the increasing of scale numbers on all three datasets. While the MS-CNN has no significant trend in performance in most cases. That is likely due to the fusion layer in MSTD-RCNN can effectively fuse the features with TD. MS-CNN uses fully connected layers to fuse features. While MSTD-RCNN uses a GRU to fuse features. Secondly, with sample input, the MSTD-RCNN has a higher level of accuracy than MS-CNN on all three datasets. The fusion layer can capture the temporary dependency which is very important for time-series classification. We can conclude that MSTD-RCNN is more effective than MS-CNN to capture the TD.  

Fig.~\ref{fig2}.b shows the F1 results on three datasets. There are similar observations as the Fig.~\ref{fig2}.a.

\subsection{Simulated trading}
The ultimate goal of financial time-series classification is to make a profit. To estimate the models' profitability, we use a simulated trading algorithm (Equation~\ref{eq19}) to evaluate these models based on their predictions on the testing sets. Table~\ref{tab5} lists the simulated trading results on three datasets. The profitability of these models is compared to the baseline strategy Buy \& Hold (B\&H) strategy. This B\&H suggests that buy in the security at the beginning and sell out at the end.

MSTD-RCNN achieves the highest profit on all three datasets. Especially, $61.44$, $126.06$ and $149.54$ higher than the most profitable baseline model. Note that B\&H strategy suffers losses due to the market is in a downward trend, while all models can make a profit. The results show that our model can not only be more accurate in classification but also more profitable than baseline models.

Next, the confusion matrix of our model is analyzed to find the cause of profitability improvement.
\begin{table}[ht]
	\centering
	\caption{profits of simulated trading.\label{tab5}}
	\setlength{\tabcolsep}{3.5mm}{\begin{tabular}{cccc}
			\toprule
			Strategies & SH000001 & SZ399005 & SZ399006\\\midrule
			B\&H & -233.13 & -221.48 & -557.53\\
			SVM & 1172.68 & 1177.10 & 7225.94\\
			RF & 1241.96 & 1247.25 & 7260.03\\
			TreNet & 1330.30 & 1255.96 & 7373.50\\
			FDNN& 1231.07 & 1273.94 & 7377.40\\
			SFM& 1316.16 & 1265.90 & 7459.88\\
			MS-CNN& 1358.41 & 1262.12 & 7427.14\\
			MS-RCNN & \bf{1419.85} & \bf{1400.93} & \bf{7609.42}\\
			\bottomrule
	\end{tabular} }
	\footnotesize{ \\Bold numbers indicate the best results}
\end{table}

\subsection{Confusion Matrix}
To find the reason for profitability improvement, we analyze its confusion matrix of our MSTD-RCNN on three datasets. For comparison, we also demonstrate the confusion matrix of MS-CNN. Since MS-CNN shares a similar structure with our model and it achieves almost the highest classification and profit performance among baseline models.

There are three categories of financial time-series: still(0), downward(1) and upward(2). Due to the impaction to profitability, the samples in downward and upward categories have a higher level of importance than the still ones.  As a result, the error classifying these two categories will make the model suffer from losses in the simulated trading. In contrast, the error classifying the still category to the other two categories have no harm to the model's profitability.

\begin{figure} 
	\centering
	\subfloat[\label{a}]{%
		\includegraphics[width=0.5\linewidth]{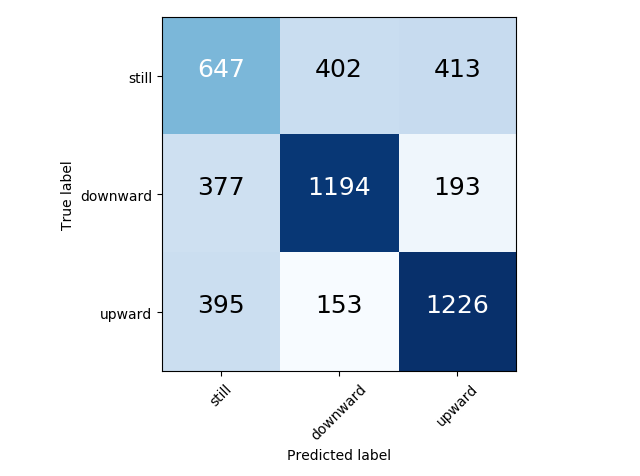}}
	\hfill
	\subfloat[\label{b}]{%
		\includegraphics[width=0.5\linewidth]{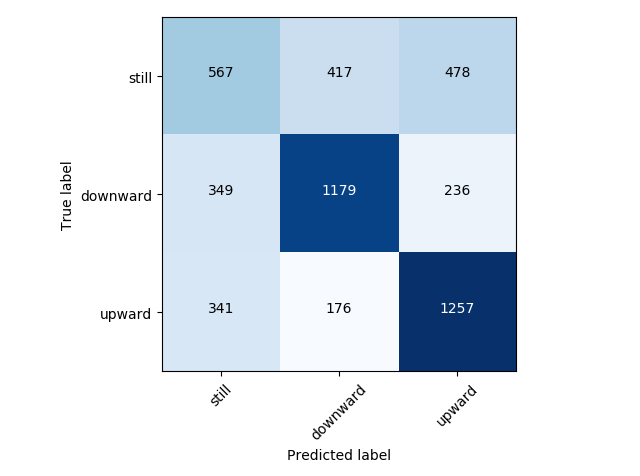}}
	\vfill
	\subfloat[\label{c}]{%
		\includegraphics[width=0.5\linewidth]{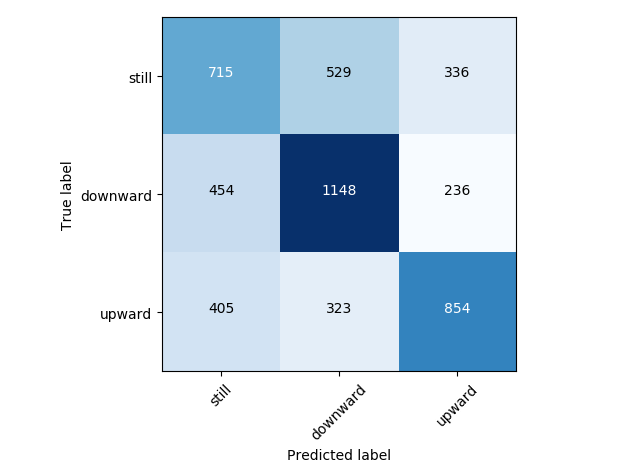}}
	\hfill
	\subfloat[\label{d}]{%
		\includegraphics[width=0.5\linewidth]{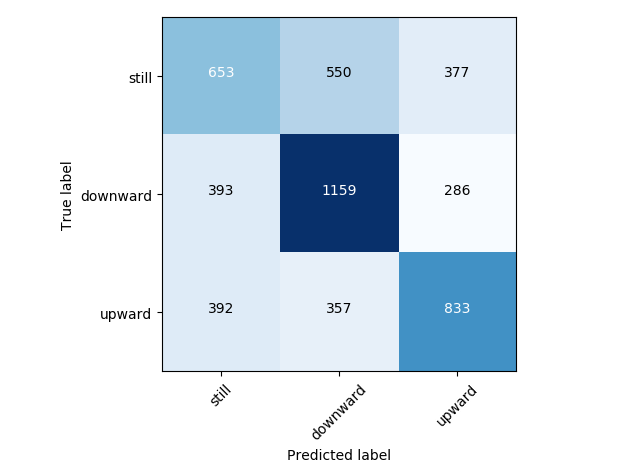}}
	\vfill
	\subfloat[\label{e}]{%
		\includegraphics[width=0.5\linewidth]{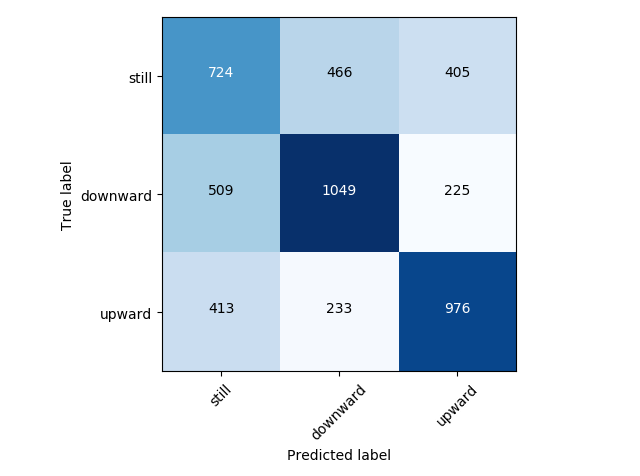}}
	\hfill
	\subfloat[\label{f}]{%
		\includegraphics[width=0.5\linewidth]{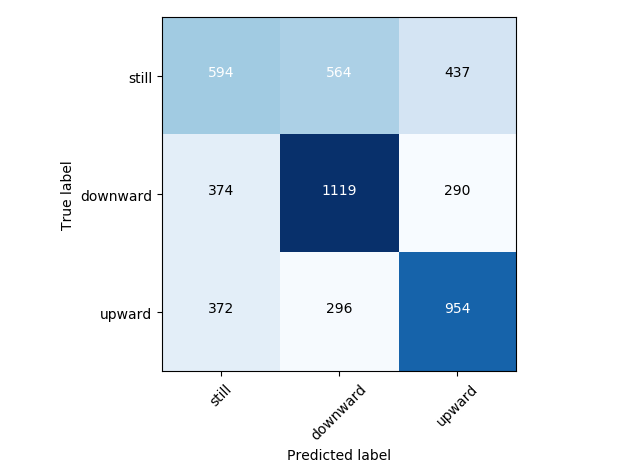}}
	\caption{(a)(c)(e) show the confusion matrix of MSTD-RCNN on SH000001, SZ399005 and SZ399006. 
		(b)(d)(f) illustrate the confusion matrix of MSTD-CNN on SH000001, SZ399005 and SZ399006.}
	\label{fig3} 
\end{figure}

Fig.~\ref{fig3} shows the confusion matrix of MSTD-RCNN and MS-CNN on SH000001, SZ399005, and SZ399006. The major observations are listed as follows:
\begin{itemize}
    \item For the error classifying "upward" to "downward", MSTD-RCNN has fewer occurrences than MS-CNN. For instance, MSTD-RCNN only error classifies $153$ sequences on SH000001, which is $23$ sequences less than MS-CNN.

	\item For the error classifying "downward" to "upward", MSTD-RCNN also has fewer occurrences than MS-CNN. For example, MSTD-RCNN error classifies $236$ sequences on SZ399005, which is $50$ sequences less than MS-CNN.
	
	\item For the "upward" and "downward", MSTD-RCNN achieves a higher level of precision than MS-CNN. Such as on SZ399006, MSTD-RCNN achieves "downward" precision $0.600$ and "upward" precision $0.608$, which are $6.2\%$ and $7.0\%$ higher than MS-CNN. 
	
\end{itemize}

MSTD-RCNN has higher upward and downward classification accuracy and lower error classifying number between upward and downward than MSTD-RCNN. Those are the accounting for that our model achieves higher profitability than MSTD-RCNN. Moreover, it is likely causing our model to achieve the highest profitability in simulated trading.

\section{Conclusion and future works}
\label{sec:conc}

This paper proposes a Multi-Scale Recurrent Convolutional Neural Network, denoted MSTD-RCNN, for financial time-series classification. The proposed method can effectively combine and utilize Multi-Scale (MS) and Temporal Dependency (TD). The convolutional units are integrated to simultaneously extract MS-features, and a GRU is used to capture the TD across multiple scales. This enables the classification of time-series with MS-property by feedforwarding a single time-scale input sequence through the network, which results in a very effective end-to-end classifier. The profitability of our model is also evaluated by a simulated trading algorithm. Extensive experimental results suggest that our MSTD-RCNN achieves state-of-the-art performance in financial time-series classification. 

In the future, we prepare to explore three potential directions to improve our MSTD-RCNN. First, different structure of feature extractors, such as the most recently Transformer~\cite{devlin2018bert} is likely an even more effective structure than CNN. Second, the attention mechanism~\cite{liu2019numerical} can be introduced to handle the long-term dependency which cannot be handled by RNN. Third, multi-source of information can be used, especially textual information such as news.

\begin{acknowledgements}
	This research was funded in part by the National Social Science Fund of China No. 2016ZDA055, and in part by the Discipline Building Plan in 111 Base No. B08004. The authors thank NVIDIA's donations of their providing GPUs used for this research. The authors would also like to thank the editor and anonymous reviewers for their precious comments on improving this article. 
\end{acknowledgements}

\bibliography{mybib.bib}
\bibliographystyle{spmpsci}

\end{document}